\documentclass[11pt,twoside]{article}

\usepackage{graphicx}
\usepackage{amssymb}
\usepackage{amsmath}
\usepackage{amsfonts}
\usepackage{fancyhdr}
\usepackage[english]{babel}
\usepackage[usenames,dvipsnames]{color}
\usepackage{palatino}
\usepackage{wrapfig}
\usepackage{float}
\usepackage{color}
\usepackage{chngcntr}
\usepackage{listings}
\usepackage{longtable}
\usepackage{multirow}
\usepackage{array}
\usepackage{rotating}
\usepackage{dirtree}
\usepackage{pxfonts}
\usepackage{acta-info}

\usepackage[dvipsnames*,svgnames]{xcolor}
\hypersetup{colorlinks=true,allcolors=DarkBlue,linkcolor=black}

\definecolor{commentgreen}{RGB}{2,112,10}
\definecolor{frenchplum}{RGB}{80,0,50}

\newcommand{\vol}{\textbf{10,} 1 (2018) 26--42}   
\newcommand{\amss}{\amssubj{%
68T45
}}     
\newcommand{\CRs}{\CRsubj{%
I.2.6
}}   
\newcommand{\keyw}{\keywords{%
Deep learning, Object recognition, Computer vision
}}

\newcommand{\singleFruitImageCount}{90380 }
\newcommand{\trainingImageCount}{67692 }
\newcommand{\testImageCount}{22688 }
\newcommand{\multiFruitImageCount}{103 }
\newcommand{\totalImageCount}{90483 }
\newcommand{\classCount}{131 }
\newcommand{\lastUpdated}{2020.05.18}

\newcolumntype{C}{>{\centering\arraybackslash}}

\begin{document}

\title{
Fruit recognition from images using deep learning
}

\maketitle

\twoauthors{%
Horea Mure\c san
}{%
\href{http://www.cs.ubbcluj.ro/}{Faculty of Mathematics and Computer Science\\ Mihail Kog\v alniceanu, 1 \\Babe\c s-Bolyai University\\Romania} 
}{%
\href{mailto:horea94@gmail.com}{horea94@gmail.com} 
}{%
\href{https://mihaioltean.github.io/}{Mihai Oltean} 
}{%
\href{http://stiinteexacte.uab.ro/}{Faculty of Exact Sciences and Engineering \\Unirii, 15-17\\"1 Decembrie 1918" University of Alba Iulia\\Romania}
}{%
\href{mailto:mihai.oltean@gmail.com}{mihai.oltean@gmail.com} 
}
\begin{abstract}                                          

In this paper we introduce a new, high-quality, dataset of images containing fruits. We also present the results of some numerical experiment for training a neural network to detect fruits. We discuss the reason why we chose to use fruits in this project by proposing a few applications that could use such classifier.

\smallskip


\end{abstract}


\section{Introduction}

The aim of this paper is to propose a new dataset of images containing popular fruits. The dataset was named Fruits-360 and can be downloaded from the addresses pointed by references \cite{fruits_360_github} and \cite{fruits_360_kaggle}. Currently (as of \lastUpdated) the set contains \totalImageCount images of \classCount fruits and vegetables and it is constantly updated with images of new fruits and vegetables as soon as the authors have accesses to them. The reader is encouraged to access the latest version of the dataset from the above indicated addresses.

Having a high-quality dataset is essential for obtaining a good classifier. Most of the existing datasets with images (see for instance the popular CIFAR dataset \cite{cifar}) contain both the object and the noisy background. This could lead to cases where changing the background will lead to the incorrect classification of the object.

As a second objective we have trained a deep neural network that is capable of identifying fruits from images. This is part of a more complex project that has the target of obtaining a classifier that can identify a much wider array of objects from images. This fits the current trend of companies working in the augmented reality field. During its annual I/O conference, Google announced \cite{lens} that is working on an application named Google Lens which will tell the user many useful information about the object toward which the phone camera is pointing. First step in creating such application is to correctly identify the objects. The software has been released later in 2017 as a feature of Google Assistant and Google Photos apps. Currently the identification of objects is based on a deep neural network \cite{google_lens_wikipedia}.

Such a network would have numerous applications across multiple domains like autonomous navigation, modeling objects, controlling processes or human-robot interactions. The area we are most interested in is creating an autonomous robot that can perform more complex tasks than a regular industrial robot. An example of this is a robot that can perform inspections on the aisles of stores in order to identify out of place items or under-stocked shelves. Furthermore, this robot could be enhanced to be able to interact with the products so that it can solve the problems on its own. Another area in which this research can provide benefits is autonomous fruit harvesting. While there are several papers on this topic already, from the best of our knowledge, they focus on few species of fruits or vegetables. In this paper we attempt to create a network that can classify a variety of species of fruit, thus making it useful in many more scenarios.

As the start of this project we chose the task of identifying fruits for several reasons. On one side, fruits have certain categories that are hard to differentiate, like the citrus genus, that contains oranges and grapefruits. Thus we want to see how well can an artificial intelligence complete the task of classifying them. Another reason is that fruits are very often found in stores, so they serve as a good starting point for the previously mentioned project.

The paper is structured as follows: in the first part we will shortly discuss a few outstanding achievements obtained using deep learning for fruits recognition, followed by a presentation of the concept of deep learning. In the second part we describe the Fruits-360 dataset: how it was created and what it contains. In the third part we will present the framework used in this project - TensorFlow\cite{tf} and the reasons we chose it. Following the framework presentation, we will detail the structure of the neural network that we used. We also describe the training and testing data used as well as the obtained performance. Finally, we will conclude with a few plans on how to improve the results of this project. Source code is listed in the Appendix.

\section{Related work}

In this section we review several previous attempts to use neural networks and deep learning for fruits recognition.

A method for recognizing and counting fruits from images in cluttered greenhouses is presented in \cite{auto_counting}. The targeted plants are peppers with fruits of complex shapes and varying colors similar to the plant canopy. The aim of the application is to locate and count green and red pepper fruits on large, dense pepper plants growing in a greenhouse. The training and validation data used in this paper consists of 28000 images of over 1000 plants and their fruits. The used method to locate and count the peppers is two-step: in the first step, the fruits are located in a single image and in a second step multiple views are combined to increase the detection rate of the fruits. The approach to find the pepper fruits in a single image is based on a combination of (1) finding points of interest, (2) applying a complex high-dimensional feature descriptor of a patch around the point of interest and (3) using a so-called bag-of-words for classifying the patch.

Paper \cite{deep_fruits} presents a novel approach for detecting fruits from images using deep neural networks. For this purpose the authors adapt a Faster Region-based convolutional network. The objective is to create a neural network that would be used by autonomous robots that can harvest fruits. The network is trained using RGB and NIR (near infra red) images. The combination of the RGB and NIR models is done in 2 separate cases: early and late fusion. Early fusion implies that the input layer has 4 channels: 3 for the RGB image and one for the NIR image. Late fusion uses 2 independently trained models that are merged by obtaining predictions from both models and averaging the results. The result is a multi modal network which obtains much better performance than the existing networks. 

On the topic of autonomous robots used for harvesting, paper \cite{orchards} shows a network trained to recognize fruits in an orchard. This is a particularly difficult task because in order to optimize operations, images that span many fruit trees must be used. In such images, the amount of fruits can be large, in the case of almonds up to 1500 fruits per image. Also, because the images are taken outside, there is a lot of variance in luminosity, fruit size, clustering and view point. Like in paper \cite{deep_fruits}, this project makes use of the Faster Region-based convolutional network, which is presented in a detailed view in paper \cite{rcnn}. Related to the automatic harvest of fruits, article \cite{auto_harvest} presents a method of detecting ripe strawberries and apples from orchards. The paper also highlights existing methods and their performance.

In \cite{robot_harvest} the authors compile a list of the available state of the art methods for harvesting with the aid of robots. They also analyze the method and propose ways to improve them.

In \cite{data_synthesis} one can see a method of generating synthetic images that are highly similar to empirical images. Specifically, this paper introduces a method for the generation of large-scale semantic segmentation datasets on a plant-part level of realistic agriculture scenes, including automated per-pixel class and depth labeling. One purpose of such synthetic dataset would be to bootstrap or pre-train computer vision models, which are fine-tuned thereafter on a smaller empirical image dataset. Similarly, in paper \cite{fruit_count} we can see a network trained on synthetic images that can count the number of fruits in images without actually detecting where they are in the image. 

Another paper, \cite{yield_prediction}, uses two back propagation neural networks trained on images with apple "Gala" variety trees in order to predict the yield for the upcoming season. For this task, four features have been extracted from images: total cross-sectional area of fruits, fruit number, total cross-section area of small fruits, and cross-sectional area of foliage.

Paper \cite{camera_angles} presents an analysis of fruit detectability in relation to the angle of the camera when the image was taken. Based on this research, it was concluded that the fruit detectability was the highest on front views and looking with a zenith angle of $60^{\circ}$ upwards. 

In papers \cite{color_texture,color_shape_feat,cucumber} we can see an approach to detecting fruits based on color, shape and texture. They highlight the difficulty of correctly classifying similar fruits of different species. They propose combining existing methods using the texture, shape and color of fruits to detect regions of interest from images. Similarly, in \cite{k_nearest_fruits} a method combining shape, size and color, texture of the fruits together with a k nearest neighbor algorithm is used to increase the accuracy of recognition.

One of the most recent works \cite{green_grape} presents an algorithm based on the improved Chan–Vese level-set model \cite{chan} and combined with the level-set idea and M-S mode \cite{mumford}. The proposed goal was to conduct night-time green grape detection. Combining the principle of the minimum circumscribed rectangle of fruit and the method of Hough straight-line detection, the picking point of the fruit stem was calculated.

\section{Deep learning}

In the area of image recognition and classification, the most successful results were obtained using artificial neural networks \cite{high_perf_conv, very_deep}. These networks form the basis for most deep learning models.

Deep learning is a class of machine learning algorithms that use multiple layers that contain nonlinear processing units \cite{overview}. Each level learns to transform its input data into a slightly more abstract and composite representation \cite{high_perf_conv}. Deep neural networks have managed to outperform other machine learning algorithms. They also achieved the first superhuman pattern recognition in certain domains \cite{superhuman_network}. This is further reinforced by the fact that deep learning is considered as an important step towards obtaining Strong AI. Secondly, deep neural networks - specifically convolutional neural networks - have been proved to obtain great results in the field of image recognition. 

In the rest of this section we will briefly describe some models of deep artificial neural networks along with some results for some related problems.

\subsection{Convolutional neural networks}
Convolutional neural networks (CNN) are part of the deep learning models. Such a network can be composed of convolutional layers, pooling layers, ReLU layers, fully connected layers and loss layers \cite{dl}. In a typical CNN architecture, each convolutional layer is followed by a Rectified Linear Unit (ReLU) layer, then a Pooling layer then one or more convolutional layer and finally one or more fully connected layer. A characteristic that sets apart the CNN from a regular neural network is taking into account the structure of the images while processing them. Note that a regular neural network converts the input in a one dimensional array which makes the trained classifier less sensitive to positional changes.

Among the best results obtained on the MNIST \cite{mnist} dataset is done by using multi-column deep neural networks. As described in paper \cite{multi_column}, they use multiple maps per layer with many layers of non-linear neurons. Even if the complexity of such networks makes them harder to train, by using graphical processors and special code written for them. The structure of the network uses winner-take-all neurons with max pooling that determine the winner neurons.

Another paper \cite{recurrent_conv} further reinforces the idea that convolutional networks have obtained better accuracy in the domain of computer vision. In paper \cite{all_conv} an all convolutional network that gains very good performance on CIFAR-10 \cite{cifar} is described in detail. The paper proposes the replacement of pooling and fully connected layers with equivalent convolutional ones. This may increase the number of parameters and adds inter-feature dependencies however it can be mitigated by using smaller convolutional layers within the network and acts as a form of regularization.

In what follows we will describe each of the layers of a CNN network.

\subsubsection{Convolutional layers}
Convolutional layers are named after the convolution operation. In mathematics convolution is an operation on two functions that produces a third function that is the modified (convoluted) version of one of the original functions. The resulting function gives in integral of the pointwise multiplication of the two functions as a function of the amount that one of the original functions is translated \cite{conv}. 

A convolutional layer consists of groups of neurons that make up kernels. The kernels have a small size but they always have the same depth as the input. The neurons from a kernel are connected to a small region of the input, called the receptive field, because it is highly inefficient to link all neurons to all previous outputs in the case of inputs of high dimensions such as images. For example, a 100 x 100 image has 10000 pixels and if the first layer has 100 neurons, it would result in 1000000 parameters. Instead of each neuron having weights for the full dimension of the input, a neuron holds weights for the dimension of the kernel input. The kernels slide across the width and height of the input, extract high level features and produce a 2 dimensional activation map. The stride at which a kernel slides is given as a parameter. The output of a convolutional layer is made by stacking the resulted activation maps which in turned is used to define the input of the next layer.

Applying a convolutional layer over an image of size 32 X 32 results in an activation map of size 28 X 28. If we apply more convolutional layers, the size will be further reduced, and, as a result the image size is drastically reduced which produces loss of information and the vanishing gradient problem. To correct this, we use padding. Padding increases the size of a input data by filling constants around input data. In most of the cases, this constant is zero so the operation is named zero padding. "Same" padding means that the output feature map has the same spatial dimensions as the input feature map. This tries to pad evenly left and right, but if the number of columns to be added is odd, it will add an extra column to the right. "Valid" padding is equivalent to no padding.

The strides causes a kernel to skip over pixels in an image and not include them in the output. The strides determines how a convolution operation works with a kernel when a larger image and more complex kernel are used. As a kernel is sliding the input, it is using the strides parameter to determine how many positions to skip.

ReLU layer, or Rectified Linear Units layer, applies the activation function max(0, x). It does not reduce the size of the network, but it increases its nonlinear properties.

\subsubsection{Pooling layers}
Pooling layers are used on one hand to reduce the spatial dimensions of the representation and to reduce the amount of computation done in the network. The other use of pooling layers is to control overfitting. The most used pooling layer has filters of size 2 x 2 with a stride 2. This effectively reduces the input to a quarter of its original size.

\subsubsection{Fully connected layers}
Fully connected layers are layers from a regular neural network. Each neuron from a fully connected layer is linked to each output of the previous layer. The operations behind a convolutional layer are the same as in a fully connected layer. Thus, it is possible to convert between the two.

\subsubsection{Loss layers}
Loss layers are used to penalize the network for deviating from the expected output. This is normally the last layer of the network. Various loss function exist: softmax is used for predicting a class from multiple disjunct classes, sigmoid cross-entropy is used for predicting multiple independent probabilities (from the [0, 1] interval).

\subsection{Recurrent neural network}
Another deep learning algorithm is the recursive neural network \cite{recurrent_conv}. The paper proposes an improvement to the popular convolutional network in the form of a recurrent convolutional network. In this kind of architecture the same set of weights is recursively applied over some data. Traditionally, recurrent networks have been used to process sequential data, handwriting or speech recognition being the most known examples. By using recurrent convolutional layers with some max pool layers in between them and a final global max pool layer at the end several advantages are obtained. Firstly, within a layer, every unit takes into account the state of units in an increasingly larger area around it. Secondly, by having recurrent layers, the depth of the network is increased without adding more parameters. Recurrent networks have shown good results in natural language processing. 

\subsection{Deep belief network}
Yet another model that is part of the deep learning algorithms is the deep belief network \cite{deep_belief}. A deep belief network is a probabilistic model composed by multiple layers of hidden units. The usages of a deep belief network are the same as the other presented networks but can also be used to pre-train a deep neural network in order to improve the initial values of the weights. This process is important because it can improve the quality of the network and can reduce training times. Deep belief networks can be combined with convolutional ones in order to obtain convolutional deep belief networks which exploit the advantages offered by both types of architectures.

\section{Fruits-360 data set}

In this section we describe how the data set was created and what it contains.

The images were obtained by filming the fruits while they are rotated by a motor and then extracting frames.

Fruits were planted in the shaft of a low speed motor (3 rpm) and a short movie of 20 seconds was recorded. Behind the fruits we placed a white sheet of paper as background. 

\begin{figure}[ht]
	\centering
    	\includegraphics[width=0.7\linewidth]{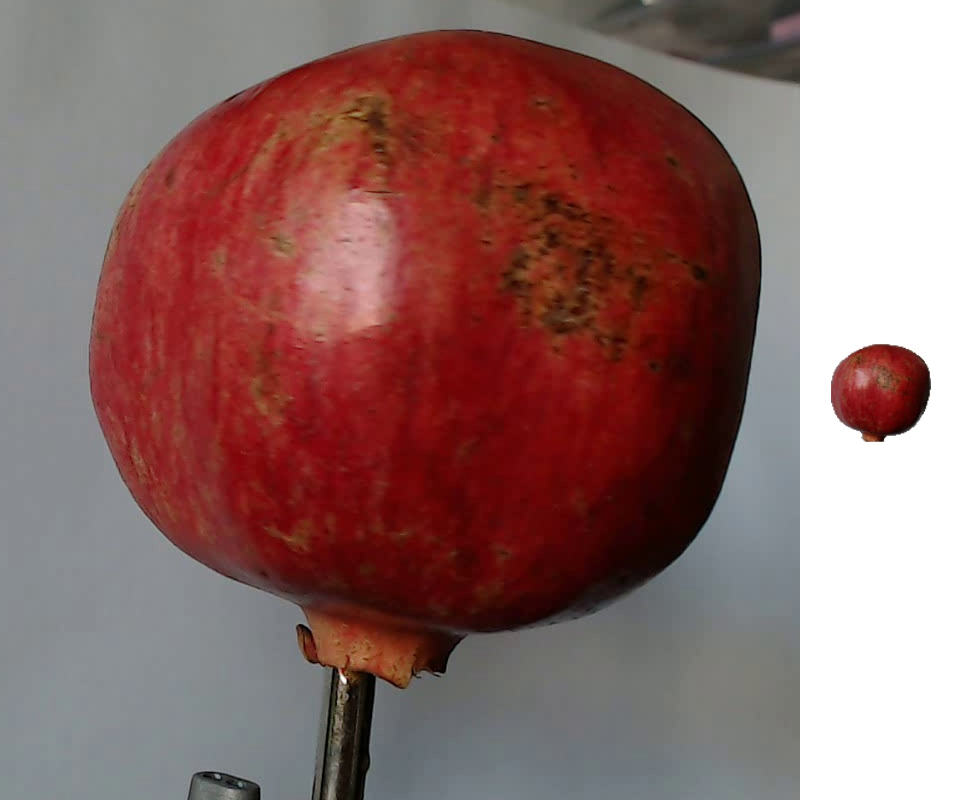}
        \caption{Left-side: original image. Notice the background and the motor shaft. Right-side: the fruit after the background removal and after it was scaled down to 100x100 pixels.}
     \label{Figure1}
\end{figure}

However due to the variations in the lighting conditions, the background was not uniform and we wrote a dedicated algorithm which extract the fruit from the background. This algorithm is of flood fill type: 
we start from each edge of the image and we mark all pixels there,
then we mark all pixels found in the neighborhood of the already marked pixels for which the distance between colors is less than a prescribed value.
we repeat the previous step until no more pixels can be marked.

All marked pixels are considered as being background (which is then filled with white) and the rest of pixels are considered as belonging to the object.
The maximum value for the distance between 2 neighbor pixels is a parameter of the algorithm and is set (by trial and error) for each movie.

Fruits were scaled to fit a 100x100 pixels image. Other datasets (like MNIST) use 28x28 images, but we feel that small size is detrimental when you have too similar objects (a red cherry looks very similar to a red apple in small images). Our future plan is to work with even larger images, but this will require much more longer training times.

To understand the complexity of background-removal process we have depicted in Figure \ref{Figure1} a fruit with its original background and after the background was removed and the fruit was scaled down to 100 x 100 pixels.

The resulted dataset has \singleFruitImageCount images of fruits and vegetables spread across \classCount labels. Each image contains a single fruit or vegetable. Separately, the dataset contains another \multiFruitImageCount images of multiple fruits. The data set is available on GitHub \cite{fruits_360_github} and Kaggle \cite{fruits_360_kaggle}.
The labels and the number of images for training are given in Table \ref{fruits_summary_table}.

\begin{longtable}{|c|c|c|}
\caption{Number of images for each fruit. There are multiple varieties of apples each of them being considered as a separate object. We did not find the scientific/popular name for each apple so we labeled with digits (e.g. apple red 1, apple red 2 etc).} 
\label{fruits_summary_table} \\

\hline \multicolumn{1}{|c|}{\textbf{Label}} & \multicolumn{1}{c|}{\textbf{Number of training images}} & \multicolumn{1}{c|}{\textbf{Number of test images}} \\ \hline 
\endfirsthead

\multicolumn{3}{c}%
{{\bfseries \tablename\ \thetable{} -- continued from previous page}} \\
\hline \multicolumn{1}{|c|}{\textbf{Label}} & \multicolumn{1}{c|}{\textbf{Number of training images}} & \multicolumn{1}{c|}{\textbf{Number of test images}} \\ \hline 
\endhead

\hline \multicolumn{3}{|r|}{{Continued on next page}} \\ \hline
\endfoot
\hline
\endlastfoot
Apple Braeburn & 492 & 164 \\ \hline
Apple Crimson Snow & 444 & 148 \\ \hline
Apple Golden 1 & 480 & 160 \\ \hline
Apple Golden 2 & 492 & 164 \\ \hline
Apple Golden 3 & 481 & 161 \\ \hline
Apple Granny Smith & 492 & 164 \\ \hline
Apple Pink Lady & 456 & 152 \\ \hline
Apple Red 1 & 492 & 164 \\ \hline
Apple Red 2 & 492 & 164 \\ \hline
Apple Red 3 & 429 & 144 \\ \hline
Apple Red Delicious & 490 & 166 \\ \hline
Apple Red Yellow 1 & 492 & 164 \\ \hline
Apple Red Yellow 2 & 672 & 219 \\ \hline
Apricot & 492 & 164 \\ \hline
Avocado & 427 & 143 \\ \hline
Avocado ripe & 491 & 166 \\ \hline
Banana & 490 & 166 \\ \hline
Banana Lady Finger & 450 & 152 \\ \hline
Banana Red & 490 & 166 \\ \hline
Beetroot & 450 & 150 \\ \hline
Blueberry & 462 & 154 \\ \hline
Cactus fruit & 490 & 166 \\ \hline
Cantaloupe 1 & 492 & 164 \\ \hline
Cantaloupe 2 & 492 & 164 \\ \hline
Carambula & 490 & 166 \\ \hline
Cauliflower & 702 & 234 \\ \hline
Cherry 1 & 492 & 164 \\ \hline
Cherry 2 & 738 & 246 \\ \hline
Cherry Rainier & 738 & 246 \\ \hline
Cherry Wax Black & 492 & 164 \\ \hline
Cherry Wax Red & 492 & 164 \\ \hline
Cherry Wax Yellow & 492 & 164 \\ \hline
Chestnut & 450 & 153 \\ \hline
Clementine & 490 & 166 \\ \hline
Cocos & 490 & 166 \\ \hline
Corn & 450 & 150 \\ \hline
Corn Husk & 462 & 154 \\ \hline
Cucumber Ripe & 392 & 130 \\ \hline
Cucumber Ripe 2 & 468 & 156 \\ \hline
Dates & 490 & 166 \\ \hline
Eggplant & 468 & 156 \\ \hline
Fig & 702 & 234 \\ \hline
Ginger Root & 297 & 99 \\ \hline
Granadilla & 490 & 166 \\ \hline
Grape Blue & 984 & 328 \\ \hline
Grape Pink & 492 & 164 \\ \hline
Grape White & 490 & 166 \\ \hline
Grape White 2 & 490 & 166 \\ \hline
Grape White 3 & 492 & 164 \\ \hline
Grape White 4 & 471 & 158 \\ \hline
Grapefruit Pink & 490 & 166 \\ \hline
Grapefruit White & 492 & 164 \\ \hline
Guava & 490 & 166 \\ \hline
Hazelnut & 464 & 157 \\ \hline
Huckleberry & 490 & 166 \\ \hline
Kaki & 490 & 166 \\ \hline
Kiwi & 466 & 156 \\ \hline
Kohlrabi & 471 & 157 \\ \hline
Kumquats & 490 & 166 \\ \hline
Lemon & 492 & 164 \\ \hline
Lemon Meyer & 490 & 166 \\ \hline
Limes & 490 & 166 \\ \hline
Lychee & 490 & 166 \\ \hline
Mandarine & 490 & 166 \\ \hline
Mango & 490 & 166 \\ \hline
Mango Red & 426 & 142 \\ \hline
Mangostan & 300 & 102 \\ \hline
Maracuja & 490 & 166 \\ \hline
Melon Piel de Sapo & 738 & 246 \\ \hline
Mulberry & 492 & 164 \\ \hline
Nectarine & 492 & 164 \\ \hline
Nectarine Flat & 480 & 160 \\ \hline
Nut Forest & 654 & 218 \\ \hline
Nut Pecan & 534 & 178 \\ \hline
Onion Red & 450 & 150 \\ \hline
Onion Red Peeled & 445 & 155 \\ \hline
Onion White & 438 & 146 \\ \hline
Orange & 479 & 160 \\ \hline
Papaya & 492 & 164 \\ \hline
Passion Fruit & 490 & 166 \\ \hline
Peach & 492 & 164 \\ \hline
Peach 2 & 738 & 246 \\ \hline
Peach Flat & 492 & 164 \\ \hline
Pear & 492 & 164 \\ \hline
Pear 2 & 696 & 232 \\ \hline
Pear Abate & 490 & 166 \\ \hline
Pear Forelle & 702 & 234 \\ \hline
Pear Kaiser & 300 & 102 \\ \hline
Pear Monster & 490 & 166 \\ \hline
Pear Red & 666 & 222 \\ \hline
Pear Stone & 711 & 237 \\ \hline
Pear Williams & 490 & 166 \\ \hline
Pepino & 490 & 166 \\ \hline
Pepper Green & 444 & 148 \\ \hline
Pepper Orange & 702 & 234 \\ \hline
Pepper Red & 666 & 222 \\ \hline
Pepper Yellow & 666 & 222 \\ \hline
Physalis & 492 & 164 \\ \hline
Physalis with Husk & 492 & 164 \\ \hline
Pineapple & 490 & 166 \\ \hline
Pineapple Mini & 493 & 163 \\ \hline
Pitahaya Red & 490 & 166 \\ \hline
Plum & 447 & 151 \\ \hline
Plum 2 & 420 & 142 \\ \hline
Plum 3 & 900 & 304 \\ \hline
Pomegranate & 492 & 164 \\ \hline
Pomelo Sweetie & 450 & 153 \\ \hline
Potato Red & 450 & 150 \\ \hline
Potato Red Washed & 453 & 151 \\ \hline
Potato Sweet & 450 & 150 \\ \hline
Potato White & 450 & 150 \\ \hline
Quince & 490 & 166 \\ \hline
Rambutan & 492 & 164 \\ \hline
Raspberry & 490 & 166 \\ \hline
Redcurrant & 492 & 164 \\ \hline
Salak & 490 & 162 \\ \hline
Strawberry & 492 & 164 \\ \hline
Strawberry Wedge & 738 & 246 \\ \hline
Tamarillo & 490 & 166 \\ \hline
Tangelo & 490 & 166 \\ \hline
Tomato 1 & 738 & 246 \\ \hline
Tomato 2 & 672 & 225 \\ \hline
Tomato 3 & 738 & 246 \\ \hline
Tomato 4 & 479 & 160 \\ \hline
Tomato Cherry Red & 492 & 164 \\ \hline
Tomato Heart & 684 & 228 \\ \hline
Tomato Maroon & 367 & 127 \\ \hline
Tomato not Ripened & 474 & 158 \\ \hline
Tomato Yellow & 459 & 153 \\ \hline
Walnut & 735 & 249 \\ \hline
Watermelon & 475 & 157 \\ \hline
\end{longtable}
\pagebreak

\section{TensorFlow library}
For the purpose of implementing, training and testing the network described in this paper we used the TensorFlow library \cite{tf}. This is an open source framework for machine learning created by Google for numerical computation using data flow graphs. Nodes in the graph represent mathematical operations, while the graph edges represent the multidimensional data arrays called tensors. 

The main components in a TensorFlow system are the client, which uses the Session interface to communicate with the master, and one or more worker processes, with each worker process responsible for arbitrating access to one or more computational devices (such as CPU cores or GPU cards) and for executing graph nodes on those devices as instructed by the master. 

TensorFlow offers some powerful features such as: it allows computation mapping to multiple machines, unlike most other similar frameworks; it has built in support for automatic gradient computation; it can partially execute subgraphs of the entire graph and it can add constraints to devices, like placing nodes on devices of a certain type, ensure that two or more objects are placed in the same space etc.

Starting with version 2.0, TensorFlow includes the features of the Keras framework\cite{keras}. Keras provides wrappers over the operations implemented in TensorFlow, greatly simplifying calls, and reducing the overall amount of code required to train and test a model.

TensorFlow is used in several projects, such as the Inception Image Classification Model \cite{inception}. This project introduced a state of the art network for classification and detection in the ImageNet Large-Scale Visual Recognition Challenge 2014. In this project the usage of the computing resources is improved by adjusting the network width and depth while keeping the computational budget constant\cite{inception}.

Another project that employs the TensorFlow framework is DeepSpeech, developed by Mozilla. It is an open source Speech-To-Text engine based on Baidu's Deep Speech architecture \cite{deep_speech}. The architecture is a state of the art  recognition system developed using end-to-end deep learning. It is simpler that other architectures and does not need hand designed components for background noise, reverberation or speaker variation.

We will present the most important utilized methods and data types from TensorFlow together with a short description for each of them.

\pagebreak
A convolutional layer is defined like this: 
\begin{lstlisting}
      Conv2D(
           no_filters,
           filter_size,
           strides,	
           padding,
           name=None
       )
\end{lstlisting}

Computes a 2D convolution over the input of shape [\textit{batch, in\_height, in\_width, in\_channels}] and a kernel tensor of shape [\textit{filter\_height, filter\_width}]. This op performs the following:

\begin{itemize}
\item Flattens the filter to a 2-D matrix with shape [\textit{filter\_height * filter\_width * in\_channels, output\_channels}]. 
\item Extracts image patches from the input tensor to form a virtual tensor of shape [\textit{batch, out\_height, out\_width, filter\_height * filter\_width * in\_channels}]. 
\item For each patch, right-multiplies the filter matrix and the image patch vector.
\item If padding is set to "same", the input is 0-padded so that the output keeps the same height and width; else, if the padding is set to "valid", the input is not 0-padded, thus the output may be smaller across the width and height
\end{itemize}

\begin{lstlisting}
        MaxPooling2D(
            filter_size,
            strides,
            padding,
            name=None
        )
\end{lstlisting}

Performs the max pooling operation on the input. \textit{filter\_size} represents size of the window over which the max function is applied. \textit{strides} represents the stride of the sliding window for each dimension of the input tensor. Similar to the Conv2D layer, the \textit{padding} parameter can be "valid"` or "same".

\begin{lstlisting}
        Activation(
            operation,
            name=None
        )
\end{lstlisting}

Computes the specified activation function given by the \textit{operation}. We are using in this project the rectified linear operation - max(features, 0).

\begin{lstlisting}
        Dropout(
            prob,
            name=None
        )
\end{lstlisting}

Randomly sets input values to 0 with probability \textit{prob}. The method scales the non zero values by 1 / \textit{1 - prob} in order to preserve the sum of the elements.

\section{The structure of the neural network used in experiments}

For this project we used a convolutional neural network. As previously described this type of network makes use of convolutional layers, pooling layers, ReLU layers, fully connected layers and loss layers. In a typical CNN architecture, each convolutional layer is followed by a Rectified Linear Unit (ReLU) layer, then a Pooling layer then one or more convolutional layer and finally one or more fully connected layer. 

Note again that a characteristic that sets apart the CNN from a regular neural network is taking into account the structure of the images while processing them. A regular neural network converts the input in a one dimensional array which makes the trained classifier less sensitive to positional changes. 

The input that we used consists of standard RGB images of size 100 x 100 pixels.

The neural network that we used in this project has the structure given in Table \ref{tab_settings}.

\begin{longtable}{|l|l|l|}
\caption{The structure of the neural network used in this paper.} 
\label{tab_settings} \\

\hline \multicolumn{1}{|c|}{\textbf{Layer type}} & \multicolumn{1}{c|}{\textbf{Dimensions}} & \multicolumn{1}{c|}{\textbf{Output}} \\ \hline 
\endfirsthead
\endhead

Convolutional        & 5 x 5 x 4           & 16           \\ \hline
Max pooling          & 2 x 2 | Stride: 2   & -            \\ \hline
Convolutional        & 5 x 5 x 16          & 32           \\ \hline
Max pooling          & 2 x 2 | Stride: 2   & -            \\ \hline
Convolutional        & 5 x 5 x 32          & 64           \\ \hline
Max pooling          & 2 x 2 | Stride: 2   & -            \\ \hline
Convolutional        & 5 x 5 x 64          & 128          \\ \hline
Max pooling          & 2 x 2 | Stride: 2   & -            \\ \hline
Fully connected      & 5 x 5 x 128         & 1024         \\ \hline
Fully connected      & 1024                & 256          \\ \hline
Softmax              & 256                 & \classCount  \\ \hline

\end{longtable}
\begin{sidewaysfigure}[htbp]
    \centering
    \includegraphics[width=\columnwidth]{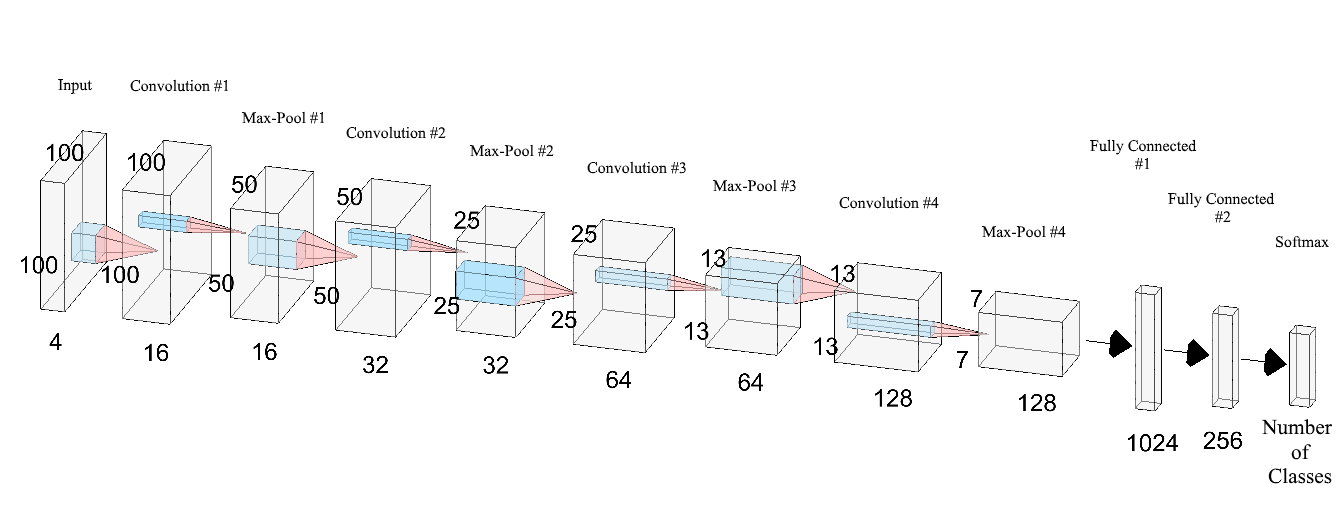}
    \caption{Graphical representation of the convolutional neural network used in experiments.}
    \label{network_structure}
\end{sidewaysfigure}

A visual representation of the neural network used is given in Figure \ref{network_structure}. 
\begin{itemize}
    \item The first layer (Convolution \#1) is a convolutional layer which applies 16 5 x 5 filters. On this layer we apply max pooling with a filter of shape 2 x 2 with stride 2 which specifies that the pooled regions do not overlap (Max-Pool \#1). This also reduces the width and height to 50 pixels each.
    \item The second convolutional (Convolution \#2) layer applies 32 5 x 5 filters which outputs 32 activation maps. We apply on this layer the same kind of max pooling(Max-Pool \#2) as on the first layer, shape 2 x 2 and stride 2.
    \item The third convolutional (Convolution \#3) layer applies 64 5 x 5 filters. Following is another max pool layer(Max-Pool \#3) of shape 2 x 2 and stride 2.
    \item The fourth convolutional (Convolution \#4) layer applies 128 5 x 5 filters after which we apply a final max pool layer (Max-Pool \#4).
    \item Because of the four max pooling layers, the dimensions of the representation have each been reduced by a factor of 16, therefore the fifth layer, which is a fully connected layer(Fully Connected \#1), has 7 x 7 x 16 inputs.
    \item This layer feeds into another fully connected layer (Fully Connected \#2) with 1024 inputs and 256 outputs.
    \item The last layer is a softmax loss layer (Softmax) with 256 inputs. The number of outputs is equal to the number of classes.
\end{itemize}

We present a short scheme containing the flow of the the training process:

\begin{lstlisting}
    epochs = 25
    
    read_images(images)
    apply_random_vertical_horizontal_flips(images)
    apply_random_hue_saturation_changes(images)
    convert_to_hsv(images)
    add_grayscale_layer(images)
    
    define_network_structure(network)
	
    for i in range(1, epochs):
		train_network(network)
	test_network(network)

\end{lstlisting}

\section{Numerical experiments}

For the experiments we used the \singleFruitImageCount images split in 2 parts: training set - which consists of \trainingImageCount images of fruits and testing set - which is made of \testImageCount images. The other \multiFruitImageCount images with multiple fruits were not used in the training and testing of the network.

Using an $ImageDataGenerator$, from the $tensorflow.keras.preprocessing.image$ package, we provide randomized input to the network. This object manages the image loading, batch generation and can perform augmentation such as vertical and horizontal flips. 

We ran multiple scenarios in which the neural network was trained using different levels of data augmentation and preprocessing:
\begin{itemize}

\item convert the input RGB images to grayscale
\item keep the input images in the RGB colorspace
\item convert the input RGB images to the HSV colorspace
\item convert the input RGB images to the HSV colorspace and to grayscale and merge them
\item apply random hue and saturation changes on the input RGB images, randomly flip them horizontally and vertically, then convert them to the HSV colorspace and to grayscale and merge them

\end{itemize}

For each scenario we used the previously described neural network which was trained for 25 epochs with batches of 50 images selected at random from the training set. Every epoch we calculated the accuracy using cross-validation. For testing we ran the trained network on the test set. The results for each case are presented in Table \ref{tab_accuracy}. All models achieved very high accuracy on the training dataset(99.98\% or above). The model trained with only RGB images obtained the best performance on the test set. A potential explanation why the model trained on augmented data performed worse than the RGB one is that the training and test images were taken into identical lighting conditions and contain the same fruit. Thus, by augmenting the images, we are introducing variation in the training set that is not found in the test set. Conversely, training on the grayscale images produces a worse result because the conversion loses all features related to color. We further studied this problem by training and testing on just the apple classes of images. The results were similar, with high accuracy on the train data, but low accuracy on the test data. 

\begin{longtable}{
|p{6cm}|Cp{2.5cm}|Cp{2.5cm}|}
\caption{Results of training the neural network on the fruits-360 dataset.} 
\label{tab_accuracy} \\

\hline \multicolumn{1}{|c|}{\textbf{Scenario}} & \textbf{Accuracy on training set} & \textbf{Accuracy on test set}\\ \hline 
\endfirsthead
\endhead

Grayscale                                       & 100\%   & 95.25\% \\ \hline %
RGB                                             & \textbf{100\%}   & \textbf{98.66\%} \\ \hline %
HSV                                             & 99.99\% & 96.09\% \\ \hline %
HSV + Grayscale                                 & 99.99\% & 96.68\% \\ \hline %
HSV + Grayscale + hue/saturation change + flips & 99.98\% & 96.44\% \\ \hline %

\end{longtable}

In order to determine the best network configuration for classifying the images in our dataset, we took multiple configurations, used the train set to train them and then calculated their accuracy on the test and training set. In Table \ref{tab_configuration_comparison} we present the results.

\begin{longtable}{|c|c|c|c|Cp{2.2cm}|Cp{2cm}|}
\caption{Results of training different network configurations on the fruits-360 dataset.} \\

\hline \textbf{Nr.} & \multicolumn{3}{|c|}{\textbf{Configuration}} & \textbf{Accuracy on training set} & \textbf{Accuracy on test set}\\ \hline
\endhead
\label{tab_configuration_comparison}

\multirow{6}{*}{1} & Convolutional & 5 x 5 & 16 & \multirow{6}{*}{100\%} & \multirow{6}{*}{98.66\%} \\ \cline{2-4}
& Convolutional & 5 x 5 & 32 & & \\ \cline{2-4}
& Convolutional & 5 x 5 & 64 & & \\ \cline{2-4}
& Convolutional & 5 x 5 & 128 & & \\ \cline{2-4}
& Fully connected & - & 1024 & & \\ \cline{2-4}
& Fully connected & - & 256 & & \\ \hline
\multirow{6}{*}{2} & Convolutional & 5 x 5 & 8 & \multirow{6}{*}{100\%} & \multirow{6}{*}{98.34\%} \\ \cline{2-4}
& Convolutional & 5 x 5 & 32 & & \\ \cline{2-4}
& Convolutional & 5 x 5 & 64 & & \\ \cline{2-4}
& Convolutional & 5 x 5 & 128 & & \\ \cline{2-4}
& Fully connected & - & 1024 & & \\ \cline{2-4}
& Fully connected & - & 256 & & \\ \hline
\multirow{6}{*}{3} & Convolutional & 5 x 5 & 32 & \multirow{6}{*}{100\%} & \multirow{6}{*}{98.41\%} \\ \cline{2-4}
& Convolutional & 5 x 5 & 32 & & \\ \cline{2-4}
& Convolutional & 5 x 5 & 64 & & \\ \cline{2-4}
& Convolutional & 5 x 5 & 128 & & \\ \cline{2-4}
& Fully connected & - & 1024 & & \\ \cline{2-4}
& Fully connected & - & 256 & & \\ \hline
\pagebreak
\multirow{6}{*}{4} & Convolutional & 5 x 5 & 16 & \multirow{6}{*}{100\%} & \multirow{6}{*}{98.35\%} \\ \cline{2-4}
& Convolutional & 5 x 5 & 16 & & \\ \cline{2-4}
& Convolutional & 5 x 5 & 64 & & \\ \cline{2-4}
& Convolutional & 5 x 5 & 128 & & \\ \cline{2-4}
& Fully connected & - & 1024 & & \\ \cline{2-4}
& Fully connected & - & 256 & & \\ \hline
\multirow{6}{*}{5} & Convolutional & 5 x 5 & 16 & \multirow{6}{*}{100\%} & \multirow{6}{*}{98.53\%} \\ \cline{2-4}
& Convolutional & 5 x 5 & 64 & & \\ \cline{2-4}
& Convolutional & 5 x 5 & 64 & & \\ \cline{2-4}
& Convolutional & 5 x 5 & 128 & & \\ \cline{2-4}
& Fully connected & - & 1024 & & \\ \cline{2-4}
& Fully connected & - & 256 & & \\ \hline
\multirow{6}{*}{6} & Convolutional & 5 x 5 & 16 & \multirow{6}{*}{100\%} & \multirow{6}{*}{97.92\%} \\ \cline{2-4}
& Convolutional & 5 x 5 & 32 & & \\ \cline{2-4}
& Convolutional & 5 x 5 & 32 & & \\ \cline{2-4}
& Convolutional & 5 x 5 & 128 & & \\ \cline{2-4}
& Fully connected & - & 1024 & & \\ \cline{2-4}
& Fully connected & - & 256 & & \\ \hline
\multirow{6}{*}{7} & Convolutional & 5 x 5 & 16 & \multirow{6}{*}{100\%} & \multirow{6}{*}{97.84\%} \\ \cline{2-4}
& Convolutional & 5 x 5 & 32 & & \\ \cline{2-4}
& Convolutional & 5 x 5 & 128 & & \\ \cline{2-4}
& Convolutional & 5 x 5 & 128 & & \\ \cline{2-4}
& Fully connected & - & 1024 & & \\ \cline{2-4}
& Fully connected & - & 256 & & \\ \hline
\multirow{6}{*}{8} & Convolutional & 5 x 5 & 16 & \multirow{6}{*}{100\%} & \multirow{6}{*}{97.95\%} \\ \cline{2-4}
& Convolutional & 5 x 5 & 32 & & \\ \cline{2-4}
& Convolutional & 5 x 5 & 64 & & \\ \cline{2-4}
& Convolutional & 5 x 5 & 64 & & \\ \cline{2-4}
& Fully connected & - & 1024 & & \\ \cline{2-4}
& Fully connected & - & 256 & & \\ \hline
\pagebreak
\multirow{6}{*}{9} & Convolutional & 5 x 5 & 16 & \multirow{6}{*}{100\%} & \multirow{6}{*}{98.17\%} \\ \cline{2-4}
& Convolutional & 5 x 5 & 32 & & \\ \cline{2-4}
& Convolutional & 5 x 5 & 64 & & \\ \cline{2-4}
& Convolutional & 5 x 5 & 128 & & \\ \cline{2-4}
& Fully connected & - & 512 & & \\ \cline{2-4}
& Fully connected & - & 256 & & \\ \hline
\multirow{6}{*}{10} & Convolutional & 5 x 5 & 16 & \multirow{6}{*}{100\%} & \multirow{6}{*}{98.25\%} \\ \cline{2-4}
& Convolutional & 5 x 5 & 32 & & \\ \cline{2-4}
& Convolutional & 5 x 5 & 64 & & \\ \cline{2-4}
& Convolutional & 5 x 5 & 128 & & \\ \cline{2-4}
& Fully connected & - & 1024 & & \\ \cline{2-4}
& Fully connected & - & 512 & & \\ \hline

\end{longtable}

From Table \ref{tab_configuration_comparison} we can see that all the tested configurations obtained perfect accuracy over the training dataset. On the test dataset, the models' performance varied slightly, with configuration nr. 1 achieving the best accuracy. Configurations nr. 6, 7 and 8 were the only ones to achieve an accuracy below 98\%.

\begin{figure}[htbp]
	\centering
    	\includegraphics[width=\textwidth]{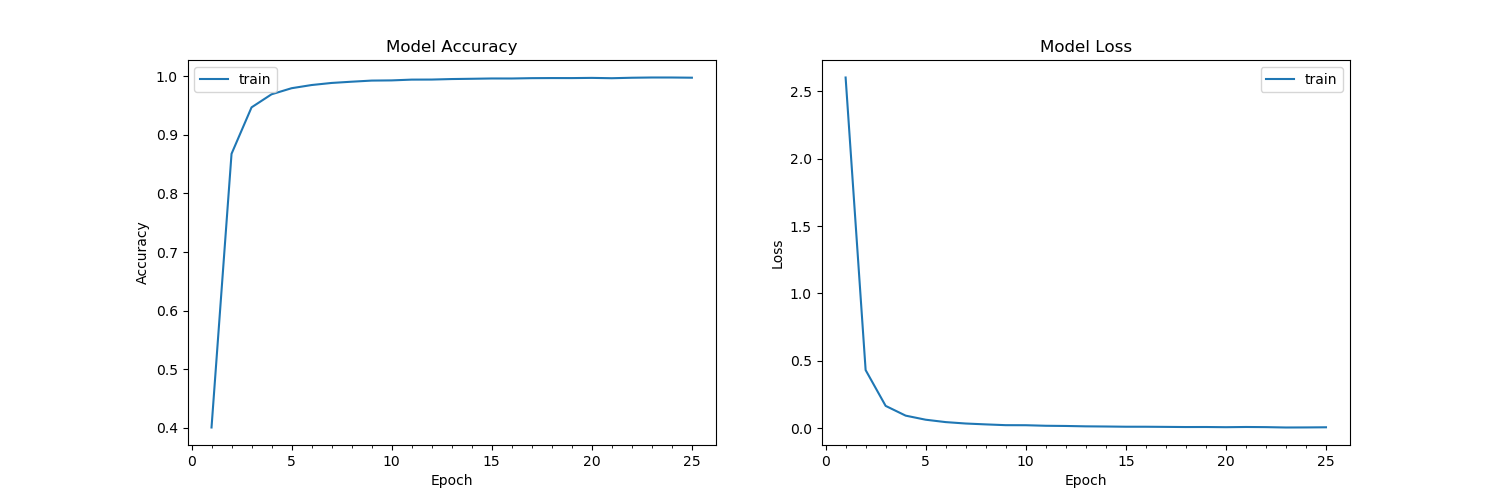}
        \caption{Accuracy and loss evolution over 25 training epochs}
     \label{accuracy_evolution}
\end{figure}

The evolution of accuracy and loss during training is given in Figure \ref{accuracy_evolution}. It can be seen that the training rapidly improves over the first 5 epochs (accuracy becomes greater than 90\%). Afterwards, the improvements are small for the rest of the epochs.

Some of the incorrectly classified images are given in Table \ref{tab_results}.

\begin{longtable}{|Cp{3cm}|Cp{3cm}|Cp{3cm}|Cp{3cm}|}
\caption{Some of the images that were classified incorrectly. On the top we have the correct class of the fruit and on the bottom we have the class (and its associated probability) that was assigned by the network.}
\label{tab_results} \\

\hline
Apple Golden 2 & Apple Golden 3 & Braeburn(Apple) & Peach \\ 
\includegraphics[width=0.2\columnwidth]{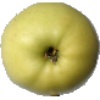} &
\includegraphics[width=0.2\columnwidth]{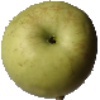} &
\includegraphics[width=0.2\columnwidth]{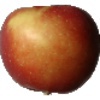} &
\includegraphics[width=0.2\columnwidth]{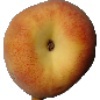} \\
Apple Golden 3 & Granny Smith (Apple) & Apple Red 2 & Apple Red Yellow \\
96.54\% & 95.22\% & 97.71\% & 97.85\%\\
\hline
Pomegranate & Peach & Pear & Pomegranate \\ 
\includegraphics[width=0.2\columnwidth]{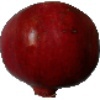} &
\includegraphics[width=0.2\columnwidth]{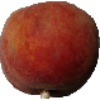} &
\includegraphics[width=0.2\columnwidth]{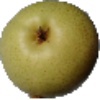} &
\includegraphics[width=0.2\columnwidth]{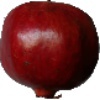} \\ 
Nectarine & Apple Red 1 & Apple Golden 2 & Braeburn(Apple) \\ 
94.64\% & 97.87\% & 98.73\% & 97.21\%\\
\hline
\end{longtable}

\section{Conclusions and further work}

We described a new and complex database of images with fruits. Also we made some numerical experiments by using TensorFlow library in order to classify the images according to their content.

From our point of view one of the main objectives for the future is to improve the accuracy of the neural network. This involves further experimenting with the structure of the network. Various tweaks and changes to any layers as well as the introduction of new layers can provide completely different results. Another option is to replace all layers with convolutional layers. This has been shown to provide some improvement over the networks that have fully connected layers in their structure. A consequence of replacing all layers with convolutional ones is that there will be an increase in the number of parameters for the network \cite{all_conv}. Another possibility is to replace the rectified linear units with exponential linear units. According to paper \cite{elu}, this reduces computational complexity and add significantly better generalization performance than rectified linear units on networks with more that 5 layers. We would like to try out these practices and also to try to find new configurations that provide interesting results.

In the near future we plan to create a mobile application which takes pictures of fruits and labels them accordingly.

Another objective is to expand the data set to include more fruits. This is a more time consuming process since we want to include items that were not used in most others related papers.

\section*{Acknowledgments}
A preliminary version of this dataset with 25 fruits was presented during the Students Communication Session from Babe\c s-Bolyai University, June 2017.

\newpage

\section*{Appendix}

In this section we present the source code and project structure used in the numerical experiment described in this paper. The source code can be downloaded from GitHub \cite{fruits_360_github}.

The source code is organized (on GitHub \cite{fruits_360_github}) as follows:
\dirtree{%
.1 root\_directory.
.2 image\_classification.
.3 Fruits-360 CNN.py.
.3 labels.txt.
.2 Test.
.2 Training.
}

In order to run the project from the command line, first make sure the PYTHONPATH system variable contains the path to the root\_directory.

Following, we will provide explanations for the code. We will begin with the definition of the general parameters and configurations of the project.

The following are defined in the \textbf{Fruits-360 CNN.py} file:
\begin{itemize}
    \item \textit{base\_dir} - the top level folder of the project
    \item \textit{test\_dir} - the folder with the Test images
    \item \textit{train\_dir} - the folder with the Training images
    \item \textit{labels\_file} - the path to the optional file that contains the used labels
    \item \textit{use\_label\_file} - boolean used to determine whether we train the network on all the classes(variable is set to false, default) or only on those specified in the \textit{labels\_file}(variable set to true)
\end{itemize}

All these configurations can be changed to suit the setup of anyone using the code.

\begin{lstlisting}
import matplotlib.pyplot as plt
import pandas as pd
import numpy as np
import seaborn as sn
import os
import tensorflow as tf
from sklearn.metrics import confusion_matrix, classification_report

from tensorflow.keras.models import Model
from tensorflow.keras.layers import Input, Dense, Conv2D, MaxPooling2D, Flatten, Activation, Dropout, Lambda
from tensorflow.keras.optimizers import Adadelta
from tensorflow.keras.preprocessing.image import ImageDataGenerator
from tensorflow.keras.callbacks import ReduceLROnPlateau, ModelCheckpoint

##############################################
learning_rate = 0.1  # initial learning rate
min_learning_rate = 0.00001  # once the learning rate reaches this value, do not decrease it further
learning_rate_reduction_factor = 0.5  # the factor used when reducing the learning rate -> learning_rate *= learning_rate_reduction_factor
patience = 3  # how many epochs to wait before reducing the learning rate when the loss plateaus
verbose = 1  # controls the amount of logging done during training and testing: 0 - none, 1 - reports metrics after each batch, 2 - reports metrics after each epoch
image_size = (100, 100)  # width and height of the used images
input_shape = (100, 100, 3)  # the expected input shape for the trained models; since the images in the Fruit-360 are 100 x 100 RGB images, this is the required input shape

use_label_file = False  # set this to true if you want load the label names from a file; uses the label_file defined below; the file should contain the names of the used labels, each label on a separate line
label_file = 'labels.txt'
base_dir = '../..'  # relative path to the Fruit-Images-Dataset folder
test_dir = os.path.join(base_dir, 'Test')
train_dir = os.path.join(base_dir, 'Training')
output_dir = 'output_files'  # root folder in which to save the the output files; the files will be under output_files/model_name
##############################################

if not os.path.exists(output_dir):
    os.makedirs(output_dir)

# if we want to train the network for a subset of the fruit classes instead of all, we can set the use_label_file to true and place in the label_file the classes we want to train for, one per line
if use_label_file:
    with open(label_file, "r") as f:
        labels = [x.strip() for x in f.readlines()]
else:
    labels = os.listdir(train_dir)
num_classes = len(labels)


# create 2 charts, one for accuracy, one for loss, to show the evolution of these two metrics during the training process
def plot_model_history(model_history, out_path=""):
    fig, axs = plt.subplots(1, 2, figsize=(15, 5))
    # summarize history for accuracy
    axs[0].plot(range(1, len(model_history.history['acc']) + 1), model_history.history['acc'])
    axs[0].set_title('Model Accuracy')
    axs[0].set_ylabel('Accuracy')
    axs[0].set_xlabel('Epoch')
    axs[0].set_xticks(np.arange(1, len(model_history.history['acc']) + 1), len(model_history.history['acc']))
    axs[0].legend(['train'], loc='best')
    # summarize history for loss
    axs[1].plot(range(1, len(model_history.history['loss']) + 1), model_history.history['loss'])
    axs[1].set_title('Model Loss')
    axs[1].set_ylabel('Loss')
    axs[1].set_xlabel('Epoch')
    axs[1].set_xticks(np.arange(1, len(model_history.history['loss']) + 1), len(model_history.history['loss']))
    axs[1].legend(['train'], loc='best')
    # save the graph in a file called "acc.png" to be available for later; the model_name is provided when creating and training a model
    if out_path:
        plt.savefig(out_path + "/acc.png")
    plt.show()


# create a confusion matrix to visually represent incorrectly classified images
def plot_confusion_matrix(y_true, y_pred, classes, out_path=""):
    cm = confusion_matrix(y_true, y_pred)
    df_cm = pd.DataFrame(cm, index=[i for i in classes], columns=[i for i in classes])
    plt.figure(figsize=(40, 40))
    ax = sn.heatmap(df_cm, annot=True, square=True, fmt="d", linewidths=.2, cbar_kws={"shrink": 0.8})
    if out_path:
        plt.savefig(out_path + "/confusion_matrix.png")  # as in the plot_model_history, the matrix is saved in a file called "model_name_confusion_matrix.png"
    return ax


# Randomly changes hue and saturation of the image to simulate variable lighting conditions
def augment_image(x):
    x = tf.image.random_saturation(x, 0.9, 1.2)
    x = tf.image.random_hue(x, 0.02)
    return x


# given the train and test folder paths and a validation to test ratio, this method creates three generators
#  - the training generator uses (100 - validation_percent) of images from the train set
#    it applies random horizontal and vertical flips for data augmentation and generates batches randomly
#  - the validation generator uses the remaining validation_percent of images from the train set
#    does not generate random batches, as the model is not trained on this data
#    the accuracy and loss are monitored using the validation data so that the learning rate can be updated if the model hits a local optimum
#  - the test generator uses the test set without any form of augmentation
#    once the training process is done, the final values of accuracy and loss are calculated on this set
def build_data_generators(train_folder, test_folder, labels=None, image_size=(100, 100), batch_size=50):
    train_datagen = ImageDataGenerator(width_shift_range=0.0, height_shift_range=0.0, zoom_range=0.0, horizontal_flip=True, vertical_flip=True, preprocessing_function=augment_image)  # augmentation is done only on the train set (and optionally validation)

    test_datagen = ImageDataGenerator()

    train_gen = train_datagen.flow_from_directory(train_folder, target_size=image_size, class_mode='sparse', batch_size=batch_size, shuffle=True, subset='training', classes=labels)
    test_gen = test_datagen.flow_from_directory(test_folder, target_size=image_size, class_mode='sparse', batch_size=batch_size, shuffle=False, subset=None, classes=labels)
    return train_gen, test_gen


# Create a custom layer that converts the original image from
# RGB to HSV and grayscale and concatenates the results
# forming in input of size 100 x 100 x 4
def convert_to_hsv_and_grayscale(x):
    hsv = tf.image.rgb_to_hsv(x)
    gray = tf.image.rgb_to_grayscale(x)
    rez = tf.concat([hsv, gray], axis=-1)
    return rez


def network(input_shape, num_classes):
    img_input = Input(shape=input_shape, name='data')
    x = Lambda(convert_to_hsv_and_grayscale)(img_input)
    x = Conv2D(16, (5, 5), strides=(1, 1), padding='same', name='conv1')(x)
    x = Activation('relu', name='conv1_relu')(x)
    x = MaxPooling2D((2, 2), strides=(2, 2), padding='valid', name='pool1')(x)
    x = Conv2D(32, (5, 5), strides=(1, 1), padding='same', name='conv2')(x)
    x = Activation('relu', name='conv2_relu')(x)
    x = MaxPooling2D((2, 2), strides=(2, 2), padding='valid', name='pool2')(x)
    x = Conv2D(64, (5, 5), strides=(1, 1), padding='same', name='conv3')(x)
    x = Activation('relu', name='conv3_relu')(x)
    x = MaxPooling2D((2, 2), strides=(2, 2), padding='valid', name='pool3')(x)
    x = Conv2D(128, (5, 5), strides=(1, 1), padding='same', name='conv4')(x)
    x = Activation('relu', name='conv4_relu')(x)
    x = MaxPooling2D((2, 2), strides=(2, 2), padding='valid', name='pool4')(x)
    x = Flatten()(x)
    x = Dense(1024, activation='relu', name='fcl1')(x)
    x = Dropout(0.2)(x)
    x = Dense(256, activation='relu', name='fcl2')(x)
    x = Dropout(0.2)(x)
    out = Dense(num_classes, activation='softmax', name='predictions')(x)
    rez = Model(inputs=img_input, outputs=out)
    return rez


# this method performs all the steps from data setup, training and testing the model and plotting the results
# the model is any trainable model; the input shape and output number of classes is dependant on the dataset used, in this case the input is 100x100 RGB images and the output is a softmax layer with 118 probabilities
# the name is used to save the classification report containing the f1 score of the model, the plots showing the loss and accuracy and the confusion matrix
# the batch size is used to determine the number of images passed through the network at once, the number of steps per epochs is derived from this as (total number of images in set // batch size) + 1
def train_and_evaluate_model(model, name="", epochs=25, batch_size=50, verbose=verbose, useCkpt=False):
    print(model.summary())
    model_out_dir = os.path.join(output_dir, name)
    if not os.path.exists(model_out_dir):
        os.makedirs(model_out_dir)
    if useCkpt:
        model.load_weights(model_out_dir + "/model.h5")

    trainGen, testGen = build_data_generators(train_dir, test_dir, labels=labels, image_size=image_size, batch_size=batch_size)
    optimizer = Adadelta(lr=learning_rate)
    model.compile(optimizer=optimizer, loss="sparse_categorical_crossentropy", metrics=["acc"])
    learning_rate_reduction = ReduceLROnPlateau(monitor='loss', patience=patience, verbose=verbose,
                                                factor=learning_rate_reduction_factor, min_lr=min_learning_rate)
    save_model = ModelCheckpoint(filepath=model_out_dir + "/model.h5", monitor='loss', verbose=verbose,
                                 save_best_only=True, save_weights_only=False, mode='min', save_freq='epoch')

    history = model.fit(trainGen, epochs=epochs, steps_per_epoch=(trainGen.n // batch_size) + 1, verbose=verbose, callbacks=[learning_rate_reduction, save_model])

    model.load_weights(model_out_dir + "/model.h5")

    trainGen.reset()
    loss_t, accuracy_t = model.evaluate(trainGen, steps=(trainGen.n // batch_size) + 1, verbose=verbose)
    loss, accuracy = model.evaluate(testGen, steps=(testGen.n // batch_size) + 1, verbose=verbose)
    print("Train: accuracy = %f  ;  loss_v = %f" % (accuracy_t, loss_t))
    print("Test: accuracy = %f  ;  loss_v = %f" % (accuracy, loss))
    plot_model_history(history, out_path=model_out_dir)
    testGen.reset()
    y_pred = model.predict(testGen, steps=(testGen.n // batch_size) + 1, verbose=verbose)
    y_true = testGen.classes[testGen.index_array]
    plot_confusion_matrix(y_true, y_pred.argmax(axis=-1), labels, out_path=model_out_dir)
    class_report = classification_report(y_true, y_pred.argmax(axis=-1), target_names=labels)

    with open(model_out_dir + "/classification_report.txt", "w") as text_file:
        text_file.write("%s" % class_report)


print(labels)
print(num_classes)
model = network(input_shape=input_shape, num_classes=num_classes)
train_and_evaluate_model(model, name="fruit-360 model")

\end{lstlisting}

\newpage
\bibliographystyle{acm}
\bibliography{Detecting_fruits_using_deep_learning}{}

\end{document}